\documentclass[journal]{IEEEtran}

\usepackage[pdftex]{graphicx}
\usepackage{epstopdf}
\usepackage{mathtools}
\usepackage{amsmath}
\usepackage{amssymb}
\usepackage{algorithm}
\usepackage{algorithmic}
\usepackage{amsfonts}
\usepackage{cite}
\usepackage{float}
\usepackage{bbold}
\usepackage{multirow}
\usepackage{tabularx}
\usepackage{threeparttable}  
\usepackage{hhline}
\usepackage{pbox}
\usepackage{pifont}

\begin{document}
\title{Algorithmic Principles of Camera-based Respiratory Motion Extraction}
\author{Wenjin~Wang,~and~Albertus~C.~den~Brinker
\thanks{W. Wang and A. C. den Brinker are with Philips Research, Eindhoven, 
The Netherlands. Correspondence: wenjin.wang@philips.com.}
}



\maketitle
\begin{abstract}
Measuring the respiratory signal from a video based on body motion has been proposed and recently matured in products for video health monitoring. The core algorithm for this measurement is the estimation of tiny chest/abdominal motions induced by respiration, and the fundamental challenge is motion sensitivity. Though prior arts reported on the validation with real human subjects, there is no thorough/rigorous benchmark to quantify the sensitivities and boundary conditions of motion-based core respiratory algorithms that measure sub-pixel displacement between video frames. In this paper, we designed a setup with a fully-controllable physical phantom to investigate the essence of core algorithms, together with a mathematical model incorporating two motion estimation strategies and three spatial representations, leading to six algorithmic combinations for respiratory signal extraction. Their promises and limitations are discussed and clarified via the phantom benchmark. The insights gained in this paper are intended to improve the understanding and applications of camera-based respiration measurement in health monitoring.
\end{abstract}


\section{Introduction}
\IEEEPARstart{C}{ameras} enable contactless respiration monitoring by measuring subtle respiratory motions from a human body (chest or abdomen). Monitoring of respiration rate is ubiquitous in a video health monitoring system, as it is one of the most important vital signs to indicate a person's health state~\cite{Bartula2013Procor}. Changes in spontaneous respiratory rate may provide early indications of physiological deterioration or delirium of a patient~\cite{brochard2012clinical}, while the average respiratory rate can provide insights into a person's well-being (e.g. sleep quality)~\cite{long2014analyzing}. Camera-based monitoring has many benefits as compared to contact-based measurement, such as capnography, electrical impedance tomography, accelerometer sensors, respiratory inductive plethysmography and structured-light plethysmography. It reduces direct mechanical contact between sensors and skin and thus eliminates the interaction with the (fragile) skin and potential infection/contamination (e.g. in COVID-19 testing) caused by contact sensors. It also increases the comfort of users and simplifies the personnel workflow, which is more suitable for long-term and continuous monitoring (24/7) as typical in clinical care units or assisted-living homes. Camera-based respiration measurement has been extensively researched in the last decade~\cite{resp2011, resp2013, resp2014, resp2015, resp2015vision, resp2018} and recently matured in products, such as baby monitoring~\cite{Jorge2017NICU}, sleep monitoring~\cite{Nochino2017Sleep}, Vital Signs App for hand-held monitoring~\cite{Philips2011VSC}, VitalEye for respiratory triggering and gating in Magnetic Resonance Imaging~\cite{Philips2018MR}, etc. 

In general, three different modalities can be used to measure the respiratory signal from a video camera: motion-based (respiratory motion at chest/abdomen)~\cite{Bartula2013Procor, Rocque2016Procor, Janssen2015Motion}, thermography-based (airflow-exchange induced temperature changes at nose/mouth)~\cite{pereira2018noncontact, Pereira2015thermal}, and photoplethysmography-based (respiration modulated blood volume changes at a peripheral living-skin)~\cite{Mirmohamadsadeghi2016Real}. Among these three principles, the motion-based modality receives most attention in research and application due to its simplicity and reproducibility, i.e. it does not require dedicated/costly thermal cameras, and has less restrictions on the characters of sensors and environment (e.g. lighting conditions) as required by camera photoplethysmography. A regular RGB/monochrome camera (e.g. webcam, IP camera or mobile phone camera) suffices as sensor for motion-based respiration measurement.

The video processing for a typical motion-based respiration monitoring system can be described as a three-step method: Region of Interest (RoI) detection, motion estimation, and respiratory signal/rate construction. We consider the second step (motion estimation) as the core of processing, while the first and last steps are front and end processes that can be achieved by leveraging generic and existing tools in video and signal processing. The primary challenge for respiratory motion estimation is the sensitivity to tiny motions, i.e. whether the algorithm is sensitive enough to capture the subtle movement induced by inhaling and exhaling at the level of sub-pixel distance. It determines the fundamental behavior of a monitoring system. We note that different from camera photoplethysmography~\cite{wang2017phd}, motion-robust respiration monitoring can hardly be designed for motion-based modality such that it is intrinsically insensitivity to non-respiratory motions (e.g. shaking of body limbs); for that other modalities with different measurement principles (e.g. thermography~\cite{Pereira2015thermal}) should be considered.
 
Regarding the core algorithms for respiratory motion extraction, various approaches~\cite{Bartula2013Procor, Janssen2015Motion} have been proposed, which are mainly based on the essence of optical flow~\cite{Lucas1981Iterative, Lucas1984Phd} and cross-correlation~\cite{Guizar2008Efficient, Sarvaiya2009Image}. These proposals have been prototyped and validated on real human subjects, including in clinical trials. However, a model for interpreting the principles of core algorithms combined with a quantifiable benchmark involving detailed control of the respiration-specific challenges is missing in the field. To address this issue, we designed a phantom study with controllable parameters and challenge factors to investigate the performance and boundary conditions for motion-based respiratory signal extraction, i.e. the phantom is created by a motor with programmable amplitude, frequency and shape for its motion signals. To explore the essence of core respiratory algorithms, we propose a model to investigate the motion estimation strategies (cross-correlation and optical flow) and spatial representations (different profiles) that are essential for this measurement. Six core algorithms are derived from the model and plugged into a fixed and automatic RoI framework, respectively, for understanding the promises and limitations of different algorithmic components.

The remainder of this paper is structured as follows. In Section II, we introduce the phantom study (setup and experimental protocol). In Sections III-IV, we introduce the model proposed for respiratory motion extraction and the core algorithms derived from the modeling. In Sections V and VI, we evaluate and discuss the core algorithms via the phantom benchmark. Finally, in Section VII, we draw the conclusions.

\begin{figure}[t!]
\centering
\includegraphics[trim=1cm 0cm 0.8cm 0cm, width=\linewidth]{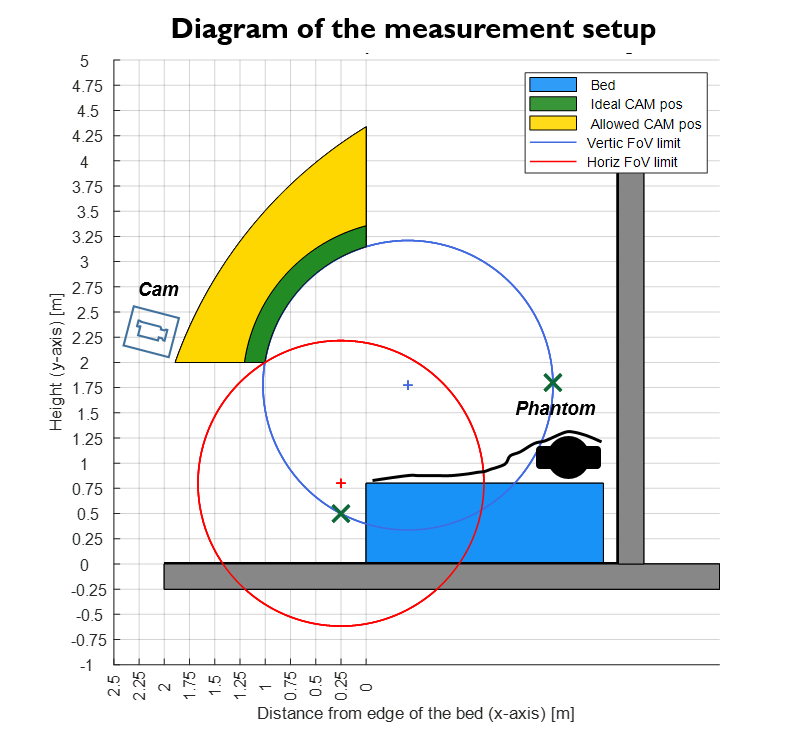}
\vspace*{-0.8cm}
\caption{The diagram of different components in the phantom setup, simulating the scenario of sleep monitoring. A programmable phantom motor, placed at the pillow side of the bed, is used to generate respiratory motion signals with controlled characteristics. The phantom is covered by a blanket to mimic a sleeping person. The green and yellow areas denote the ideal and allowed camera positions for the monitoring of the bed. The red and blue circles denote limit of field of view for the camera in horizontal and vertical directions.}
\label{fig:phantom}
\vspace*{-0.2cm}
\end{figure}

\begin{figure}[t!]
\centering

\begin{minipage}[t!]{\linewidth}
\includegraphics[trim=0cm 0cm 0cm 0cm, width=\linewidth]{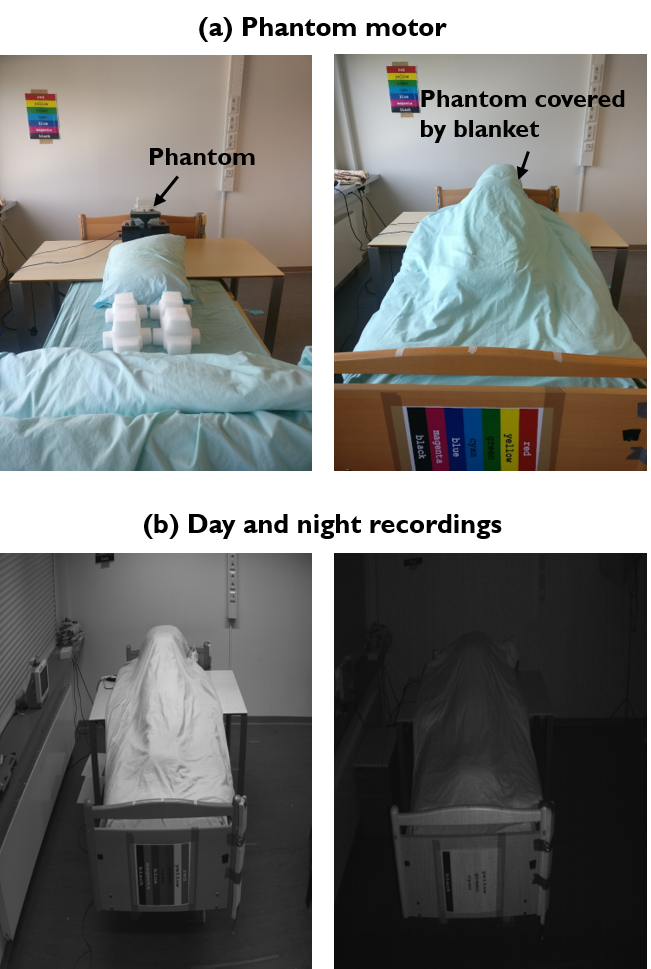}
\vspace*{-0.4cm}
\caption{(a) Illustration of the phantom setup, where a phantom motor was used to generate respiratory motion signals with controlled characteristics. It was placed at the pillow side of the bed and covered by a blanket to mimic a sleeping person. (b) Snapshots of video recordings made in the day and night categories for a visual comparison.}
\label{fig:setup}
\end{minipage}
\vspace*{0.4cm}

\begin{minipage}[t!]{\linewidth}
\centering
\includegraphics[trim=0.2cm 0cm 0cm 0cm, width=\linewidth]{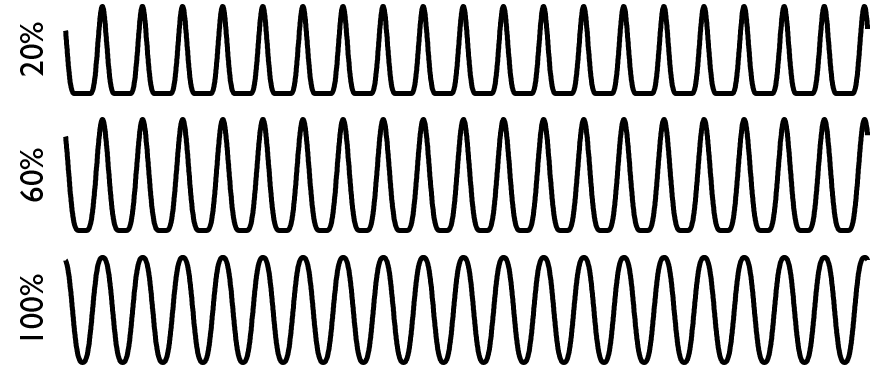}
\vspace*{-0.6cm}
\caption{The phantom signals generated with different duty circles (20\%, 60\% and 100\%). The major difference caused by duty circles is in the signal waveform morphology.}
\label{fig:duty_circle}
\vspace*{-0.2cm}
\end{minipage}

\end{figure}

\section{Phantom setup and measurement}

\subsection{Phantom setup}
The phantom study was conducted in a lab environment that simulates the scenario of sleep monitoring. Fig.~\ref{fig:phantom} illustrates the setup: a physical phantom model that generates respiration-like motion signals was positioned on the upper part of the bed and covered by a blanket to mimic a sleeping person. A camera was mounted on a tripod that is 1.8\,m in front of the bed with 2\,m height to record the scene.

For the camera sensor, we used the ON Semiconductor MT9P006, which is a 1/2.5-inch CMOS active-pixel digital image sensor. It features the low-noise CMOS imaging technology that achieves near-CCD image quality (based on Signal-to-Noise Ratio (SNR) and low-light sensitivity) while maintaining the inherent size, cost, and integration advantages of CMOS. When making video recordings, all auto-adjustment functions of the camera (e.g. auto-focus, auto-gain, auto-white-balance, auto-exposure) were switched off. The videos were recorded in an uncompressed monochrome format (with $480\times360$\,pixels, 8\,bit depth) at a constant frame rate of 15 frames per second (fps). The relatively-low image resolution is intended for investigating the challenging conditions of respiratory monitoring in practical settings with embedded devices and implementation (e.g. products based on edge computing).

The study is aimed at investigating and quantifying the motion sensitivity of core respiratory algorithms. We used a programmable motor to generate phantom motion signals where the signal characteristics (e.g. amplitude, frequency and shape) can be controlled and quantified. A broad range of breathing amplitudes and frequencies (e.g. shallow to deep and slow to fast breathing) were simulated, considering adult and neonatal breathing. A realistic setting for the minimum chest/abdominal excursion in spontaneous breathing was not determined, though \cite{adedoyin2012reference} reported that a thoracic chest expansion can be as low as 2\,mm.  

\begin{table}[!t]
\centering
\renewcommand{\arraystretch}{1.3}
\label{tab:4}
\caption{Recording protocol for each video session (day category 2 mm as an example).}
\vspace{-0.2cm}
\begin{tabular}{c|ccccc}
\hline
\hline
Time & Duration & Frequency & Duty circle & Level & Category\\
\hline
00:00:00 & 150 sec & 60 bpm & 10\% & 2 mm & Day\\
00:02:30 & 150 sec & 60 bpm & 30\% & 2 mm & Day\\
00:05:00 & 150 sec & 60 bpm & 50\% & 2 mm & Day\\
\hline
00:07:30 & 150 sec & 5 bpm & 10\% & 2 mm & Day\\
00:10:00 & 150 sec & 5 bpm & 30\% & 2 mm & Day\\
00:12:30 & 150 sec & 5 bpm & 50\% & 2 mm & Day\\
\hline
00:15:00 & 150 sec & 12 bpm & 10\% & 2 mm & Day\\
00:17:30 & 150 sec & 12 bpm & 30\% & 2 mm & Day\\
00:20:00 & 150 sec & 12 bpm & 50\% & 2 mm & Day\\
\hline
00:22:30 & 150 sec & 8 bpm & 10\% & 2 mm & Day\\
00:25:00 & 150 sec & 8 bpm & 30\% & 2 mm & Day\\
00:27:30 & 150 sec & 8 bpm & 50\% & 2 mm & Day\\
\hline
00:30:00 & 150 sec & 20 bpm & 10\% & 2 mm & Day\\
00:32:30 & 150 sec & 20 bpm & 30\% & 2 mm & Day\\
00:35:00 & 150 sec & 20 bpm & 50\% & 2 mm & Day\\
\hline
00:37:30 & 150 sec & 40 bpm & 10\% & 2 mm & Day\\
00:40:00 & 150 sec & 40 bpm & 30\% & 2 mm & Day\\
00:42:30 & 150 sec & 40 bpm & 50\% & 2 mm & Day\\
\hline
\hline
\end{tabular}
\vspace{-0.2cm}
\end{table}

\subsection{Measurement protocol}

It is expected that the core respiratory algorithms have more difficulties to deal with the dark/low-light conditions in view of camera sensor noise. We therefore set two lighting categories: day and night, and simulate different breathing amplitudes per category:
\begin{itemize} 
	\item Day: [0.5 1 2 3 4 5 6] mm
	\item Night: [2 3 4 5 6 7 8] mm
\end{itemize}
For night-time processing, we set the minimum excursion to 2\,mm, as suggested in the literature. As an extra challenge, we included even less excursion for the day-time recordings to explore the boundaries, i.e. the minimum breathing amplitude is set to 0.5\,mm. We mention that since the phantom motor is covered by a blanket (thick textile layer, see Fig.~\ref{fig:setup}~(a)), the motion strength that can be perceived by the camera was further reduced. An example of day and night recording is shown in Fig.~\ref{fig:setup}~(b). We expect a lower performance in cases of low breathing amplitudes and less illumination (night time).

For each video recording (each breathing amplitude), we further program the phantom to allow variations of other signal characteristics such as breathing frequency and duty circle:
\begin{itemize} 
	\item Frequency: [5 8 12 20 40 60] breath per minute (bpm)
	\item Duty cycle: [20 60 100]\%
\end{itemize}
The duty circle is related to the signal waveform morphology (see Fig.~\ref{fig:duty_circle}). Each cycle consists of two signal pieces: a raised cosine $1+\cos(\alpha)$ with $\alpha$ running from $-\pi$ to $\pi$ and a zero signal. The percentage of the non-zero signal is called the duty cycle. The lower duty cycles (20\%) are more characteristic for typical breathing cycles than pure sinusoidal behavior. The lower duty circles are expected to be more challenging as the flat/round signal valleys have less motion information. We also expect that breathing frequency is highly relevant for motion sensitivity, i.e. given two video frames with fixed time delay, the displacement generated with high breathing frequency will be more significant than that with low breathing frequency. 

To summarize, we created two lighting categories (day and night) and defined seven breathing amplitudes per category to investigate the motion sensitivity. Each breathing amplitude has an independent video recording session. For each session, we modulated the breathing frequencies (six) and duty circles (three) to increase the variety of phantom signal characteristics. The complete recording protocol for one video session (e.g. day 2\,mm) is shown in Table 1. Each session has a duration of 45 minutes based on the protocol, thus the total length of the benchmark dataset (14 sessions) is 630 minutes (10.5 hours). 



\section{Mathematical model}
In this section, we propose the following model for respiratory motion extraction. A camera is creating a sampled version of the light intensity at its sensor surface $C(x,y,t)$, with two spatial coordinates $x$ and $y$ and time $t$, all three coordinates are taken here as continuous variables. This intensity profile is described as the product of two processes: the illumination strength $I$ and the properties of the reflecting material. We assume that there is a material in the focal plane of the camera that is not dominated by specular reflection. Certainly for Near-Infrared (NIR) camera this is the common situation. The reflectance is denoted as $R(x,y,t)$.

\begin{figure}[t!]
\centering
\includegraphics[trim=0cm 0cm 0cm 0cm, width=\linewidth]{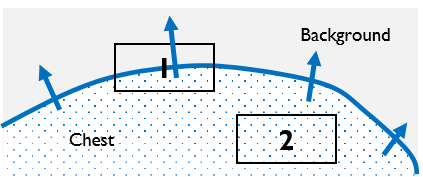}
\vspace*{-0.4cm}
 \caption{Schematic presentation of an image consisting of a moving chest against a background and two blocks in the image.}
 \label{fig:model}
\end{figure}

Consider the situation in Fig.~\ref{fig:model}. We have an image consisting of two parts: a chest (lower left part)  in front 
of the background (top right) separated by a moving boundary (curved line). The chest is expanding and contracting due to inhaling and exhaling, and its movement is not in the same direction everywhere, nor with the same amplitude. We consider two small blocks in this image. Block~1 contains part of the boundary between background and foreground and we can assume a sharp transition (as the chest is assumed in focus). In Block~2 there is no boundary between fore- and background, but the monitored chest area may have a 
pattern from which movement can be inferred. It is obvious that movement can be inferred only if the reflectance is not homogeneous in the motion direction.

For Block~2, we make the assumption of uniform movement of entire block. For Block~1, we make the assumption that the movement of the boundary and chest area inside the block is uniform. In case the background has a uniform intensity (no pattern), the two assumptions can be considered equal as an object with a uniform intensity could be considered to move with arbitrary velocity (no pattern changes).

For short time intervals we assume that the speed is constant and the camera signal is approximated as
 \begin{equation}
 \begin{split}
  C(x,y,t) & =  I(x,y,t) R(x,y,t) \\
             & =  I(x,y,t) P(x-v_xt,y-v_yt),
 \end{split}
 \label{EqMod1}
 \end{equation}
where $P$ is a 2D (shifting) pattern, $v_x$ and $v_y$ are the velocities in $x$ and $y$-direction, respectively, inside the block. Apart from a displacement, the motion is assumed to have no impact on the reflectance. This also means that movements in the direction to or from the camera, thereby leading to an intensity change, are ignored. There is only motion inside the focal plane, and the block size is small and observed time interval is short such that, locally, uniform motion can be assumed.

Ideally, one would like to have a constant uniform illumination $I(x,y,t) = I_0$. In practice, this is not the case as the scene is typically (directly or indirectly) illuminated by various sources, where some of the sources may be modulated by moving obstacles, like shifting clouds, fluttering curtains, etc. We do not consider these cases of (moving) shadows induced by changing optical pathways in an indoor monitoring condition. For short time intervals we assume that the light is spatially uniform yet may be modulated in time such that the illumination can be described as:
 \begin{equation}
  I(x,y,t) =  I_0 (1+ I_m(t)),
  \label{EqIllu}
 \end{equation}
where $I_m$ is a temporal modulation pattern. Substituting (\ref{EqIllu}) in (\ref{EqMod1}) and describing $P$
as a steady and modulating zero-mean part as  $P(\alpha,\beta) = P_0 (1+ P_m(\alpha,\beta))$ leads to the following approximation:
 \begin{equation}
  C(x,y,t) 
  \approx I_0 P_0 \{1+I_m(t) \}  \{1 + P_m(x-v_xt,y-v_yt) \}. 
  \label{EqMod2}
 \end{equation}
In total, the model tells us to expect four parts: a steady DC component, a temporal modulation, a moving pattern, and an interaction between light source modulation and the moving pattern. 

The image $C$ is the input to the camera, being sampled and quantized. This is accounted for an additional noise source: 
\begin{equation}
  C(x,y,t) 
  \approx I_0 P_0   \{1 + P_m(x-v_xt,y-v_yt) \}  + N(x,y,t),
  \label{EqFullModel}
 \end{equation}
where $N$ can be used to account for unequal pixel offsets and signal quantization. For simplicity, we assume in the remainder that we have zero-mean noise signals that are uncorrelated over time and space and of equal strength:
 \begin{equation}
 \label{EqZeroMeanNoise}
  \begin{split}
 \scriptstyle{\cal{E}} \{N(x,y,t)\} & =   \scriptstyle 0,\\ 
  \scriptstyle{\cal{E}} \{N(x_1,y_1,t_1)N(x_2,y_2,t_2)\} & =  \scriptstyle \sigma_N^2 \delta(x_1-x_2)\delta(y_1-y_2)\delta(t_1-t_2),
  \end{split}
 \end{equation}
where ${\cal E\{\cdot\}}$ denotes the expectation operator and $\delta(\cdot)$ the Dirac delta function.

Essentially, we are only interested in tracking the movement pattern $v_x$, $v_y$ contained in $P_m$.
Qualitatively, the character of the motion is that it is quasi-periodic and quantitatively it is limited to
a certain repetitions rates and small movements (as discussed before). Furthermore, we may assume that the camera is focused and therefore the boundary in Block 1 will be sharp. Similarly any clear boundary in Block 2 will be present in $C$.

\section{Motion extraction strategies}

Before any further processing, it is advantageous to remove non-informative content. Due to (\ref{EqZeroMeanNoise}), we have:
\begin{equation}
  \overline{C}(t) =  I_0 P_0 \{1+I_m(t) \}),
  \label{EqSpatialAve}
 \end{equation}
where  $\bar{\ }$ on top of a variable indicates the average over the spatial dimensions.
We define $\widetilde{C}(x,y,t)$ as the DC normalized camera signal:
 \begin{equation}
 \begin{split}
  \widetilde{C}(x,y,t) & =  {C}(x,y,t)/ \overline{C}(t) \\
  				  & =  P_m(x-v_xt,y-v_yt) + N(x,y,t)/\overline{C}(t).
   \end{split}
  \label{EqFullModelReduced}
 \end{equation}
The signal $\widetilde{C}$ contains the desired information (the motion vector ($v_x$, $v_y$) in the signal $P_m$) plus
a noise signal with a time-varying strength. In the following subsections, we highlight two commonly used motion strategies (cross-correlation and optical flow), with a particular focus on how to use them to address the model, i.e. the obtained insights are specific to respiratory motion estimation.

\begin{figure*}[t!]
\centering
\includegraphics[trim=0cm 0cm 0cm 0cm, width=\linewidth]{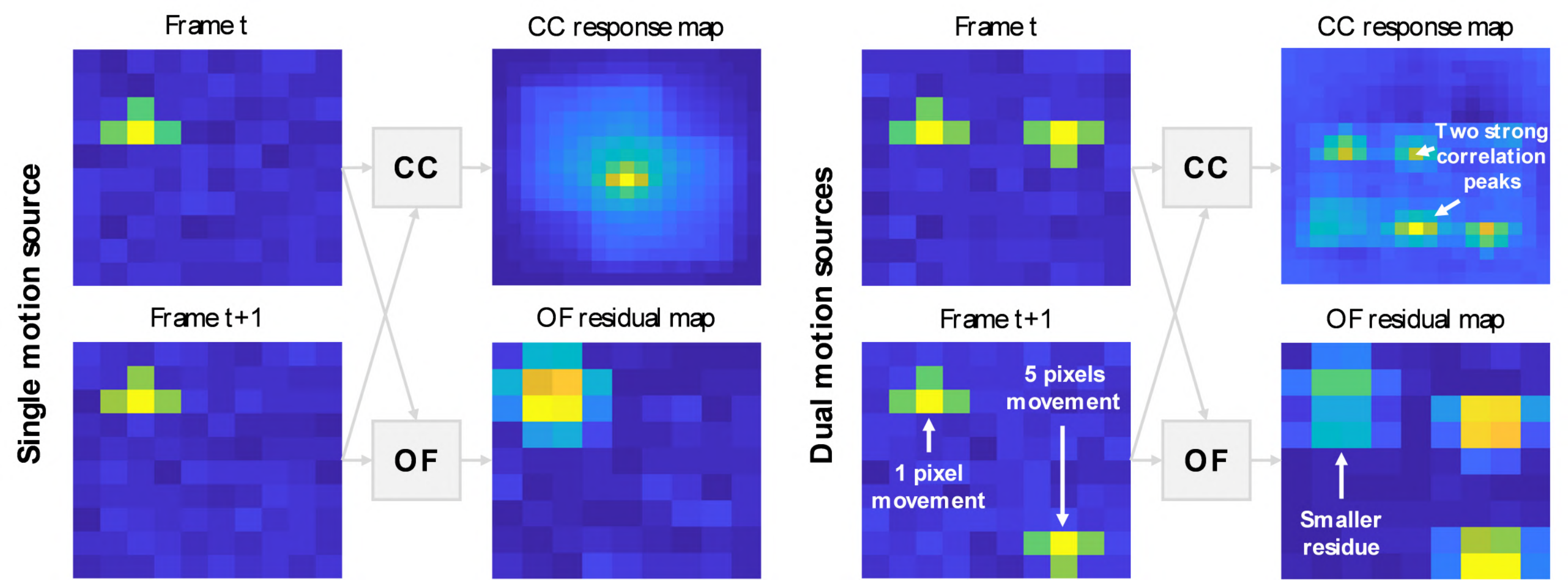}
\vspace*{-0.6cm}
\caption{Illustration of cross-correlation (CC) and optical flow (OF) in addressing small and large motions. The fundamentals of CC (e.g. correlation map) and OF (e.g. residual error map of regression) are shown. If two motion sources appear simultaneously, CC handles them equally. It shows two strong correlation response peaks that may confuse the peak selection. OF can bias to the motion with smaller amplitude due to the use of small kernels for spatial gradients computation, i.e. the large motion shows larger residual errors during the regression of using small kernels.}
\label{fig:cc_of}
\vspace*{-0.2cm}
\end{figure*}

\subsection{Cross-correlation}

We denote the 2D auto-correlation function associated with $P_m$ as $A_P(\alpha,\beta)$. Its maximum occurs at 
$\alpha=\beta=0$. Determining the 2D spatial cross-correlation function $G$ of the $\widetilde{C}$
at $t_1$ and $t_2$ gives a shifted version of the auto-correlation function:
 \begin{equation}
 G(\alpha,\beta) = A_P(\alpha - v_x (t_2-t_1),\beta - v_y (t_2-t_1)).
 \end{equation}
As terms due to $N$ average out, determining the position of the maximum of $G$ and knowing $\Delta t = t_2-t_1$ allows us to calculate the motion. Formally, 
 \begin{eqnarray}
  \hat{\alpha}, \hat{\beta}  & = & \arg \max_{\alpha,\beta} G,\\
  v_x & = & \hat{\alpha}/\Delta t, \\
  v_y & = & \hat{\beta}/\Delta t.
   \label{CC}
 \end{eqnarray}
In practice, an RoI is taken encompassing the chest area and $\Delta t$ is taken as two consecutive frames. There is however much more freedom. To enable the sub-pixel motion estimation, $G$ measured on a coarse pixel-level grid can be interpolated (e.g. by linear interpolation) to a sub-pixel level, i.e. with an accuracy of 0.01 pixel before determining the maximum.

In view of Fig.~\ref{fig:model} featuring unequal motion (directions and strengths) at different parts, it may be advantageous to use multiple segments to determine the local motion and at a later stage determine how these velocities will be combined into a single respiratory signal. For example, in a larger region, the total velocity strength could be determined (i.e. ignoring orientation) and a 
respiratory signal can be created as an average of these velocity signals. In case of the shallow breathing, it may be advantageous to consider not or not only adjacent frames, but to use spatial correlation over multiple frames with a longer time interval.

The standard approach uses typically larger areas to retrieve the signal for cross-correlation. In such an approach, we foresee two fundamental limitations in cross-correlation regarding the issue of ``motion sensitivity'': (i) the correlation is firstly estimated on the original pixel level and then interpolated to the sub-pixel level. The sub-pixel shift is created from interpolation rather than direct measurement; (ii) the correlation uses a relatively large aperture or receptive field (i.e. an image block, can hardly be $2\times2$\,pixels), so the performance is dependent on the structure/texture of the profile. If a portion of the block is static (non-respiratory region like background), it may stabilize the registration and reduce the sensitivity to the moving part, especially when the static part has more textures than the moving part.
 
\subsection{Optical flow}
For a single moving pattern in the image $\widetilde{C}(x,y,t) = P_m(x-v_xt,y-v_yt)$, the velocity can also be determined by the optical flow. Defining $\alpha = x-v_xt$ and $\beta =y-v_yt$, we have partial derivatives of $\widetilde{C}$ as:
 \begin{eqnarray}
 \frac{\partial \widetilde{C}}{\partial x} & = &  \frac{\partial P_m}{\partial \alpha} \frac{d\alpha}{dx} ,\\
 \frac{\partial \widetilde{C}}{\partial y} & = &  \frac{\partial P_m}{\partial \beta} \frac{d\beta}{dy} ,\\
 \frac{\partial \widetilde{C}}{\partial t} & = &  \frac{\partial P_m}{\partial \alpha} \frac{d\alpha}{dt} + \frac{\partial P_m}{\partial \beta} \frac{d\beta}{dt} \, .
 \end{eqnarray}
Combining the above gives:
 \begin{equation}
  \label{OF}
 \frac{\partial \widetilde{C}}{\partial t} = -v_x  \frac{\partial \widetilde{C}}{\partial x} -v_y \frac{\partial \widetilde{C}}{\partial y},
 \end{equation}
showing that the (local) velocities are related to the derivatives in time and space. In case of an extra noise term $N$,
the noise will translate to $N$ on all estimates for partial derivatives. Considering discrete-time pixels from a digital camera sensor, there are multiple ways to obtain estimates of these partial derivatives, and each has its specific error-propagation. According to~\cite{Lucas1981Iterative}, $(\frac{\partial \widetilde{C}}{\partial x}, \frac{\partial \widetilde{C}}{\partial y})$ can be approximated by image spatial gradients, obtained by the convolution with high-frequency kernels that compute gradients on horizontal and vertical directions, e.g. $[-1,1]$ and $[-1,1]^\top$; $\frac{\partial \widetilde{C}}{\partial t}$ can be approximated by image temporal gradients like the subtraction of two video frames. The spatial gradient aperture is defined by the kernel size, while the temporal gradient aperture is defined by the latency between two frames. Our hypothesis is that the small kernel size is essential for attaining high sensitivity to small motions. These will be discussed in details in the experimental section. 

\begin{figure*}[t!]
\centering
\includegraphics[trim=0cm 0cm 0cm 0cm, width=\linewidth]{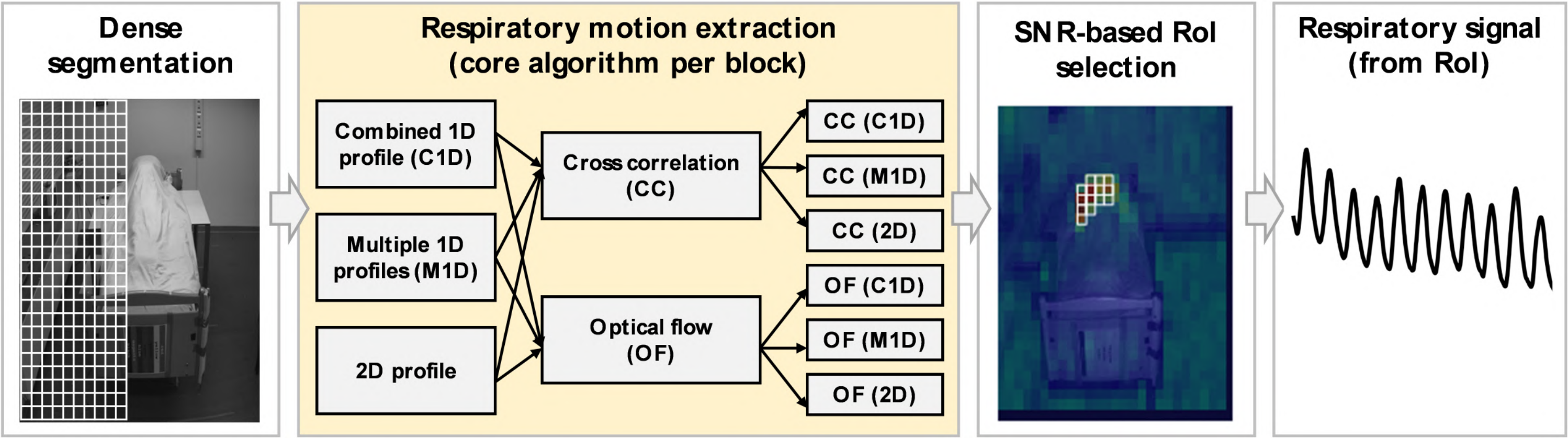}
\vspace*{-0.6cm}
\caption{The flowchart of the auto-RoI framework for end-to-end respiratory signal extraction. It contains three main steps: dense segmentation, respiratory motion extraction (with six different core algorithms), and SNR-based RoI selection. The second step is the focus of this paper.}
\label{fig:autoroi}
\vspace*{-0.2cm}
\end{figure*}

For both the 1D and 2D cases, a single measurement of the partial derivatives is insufficient (i.e. under-determined system). By assuming the same motion in multiple pixels in a block, we can create an over-determined system of (\ref{OF}) and search for least-squares solution, which is a method also beneficial in view of the additive sensor noise (see (\ref{EqFullModel})). Again, similar to cross-correlation, it may be advantageous to determine the velocities in small blocks in order to avoid information spreading over areas with different motion directions and strengths. This is the standard way of operation, and also the approach deployed in the benchmark.


%

The discussions above highlight three essential differences between optical flow and cross-correlation: (i) optical flow has a much smaller aperture or receptive field for spatial processing. Its spatial gradients only consider the neighboring pixels in a small (derivative) kernel, which is therefore more sensitive to local spontaneous changes that are relevant for small motions. If multiple motion sources occur in one block, it may bias to the ones with smaller amplitudes, which is a property preferred for respiratory motion extraction as it is vital as compared to other body motions (see Fig.~\ref{fig:cc_of}); (ii) the least-squares regression in optical flow is an implicit way of pruning outliers; and (iii) the regression is performed on spatial and temporal derivatives, making it less sensitive to biases due to the presence of static texture.


\subsection{Spatial representations}
Different spatial representations (e.g. profiles) were used as input for the discussed motion estimation strategies. The profile characterizes the structure/texture of an image patch. Most studies~\cite{Bartula2013Procor, Janssen2015Motion} assumed that vertical direction contains most respiratory energy in the target scenario, and thus only estimate the vertical motion ($v_y$ in (\ref{CC}) and (\ref{OF})) to derive the respiratory signal. In our case, we use three different spatial representations, of which two are 1D and one is 2D. All our 1D profiles are vertically oriented and thus in line with previous studies.

\subsubsection{Combined 1D profile (C1D)} It combines all the image pixels in a patch on the horizontal direction to generate a vertical vector (e.g. the approach used by ProCor~\cite{Bartula2013Procor,Rocque2016Procor}). This process is done by averaging the pixels over the horizontal direction, essentially assuming rigid motion in only the vertical direction. The benefits are less computations for motion estimation and lower sensor noise for the combined 1D profile. The drawback is that it may fuse moving pixels and stationary pixels or the pixels with different motion sources, which decreases the sensitivity of local motion estimation or even pollutes the measurement.

\subsubsection{Multiple 1D profiles (M1D)} It treats each image column as an independent vertical profile, thus one patch has multiple 1D profiles. The benefit is that it preserves the sensitivity of local motion estimation on the vertical direction, i.e. stationary and moving profiles in a patch can be analyzed separately. Though the sensor noise per profile is larger than the combined 1D profile, it can be reduced by combining multiple vertical shifts estimated from the profiles afterwards. The drawback of this approach that it does not consider the horizontal motion, i.e. horizontal motions may cause mismatching of column profiles. 

\subsubsection{2D profile} The third approach is to consider the whole 2D image patch as a single entity to estimate the 2D displacement, though only the vertical shift will be used later. The benefit is that it can use one more degree of freedom (horizontal matching) to improve the accuracy of registration of profiles, while the drawback is obviously the increased computations as compared to the combined or multiple 1D profiles.

With two motion estimation strategies (cross-correlation (CC) and optical flow (OF)) and three spatial profiles (C1D, M1D, 2D), we create six different combinations of core algorithms for respiratory motion extraction, namely CC-C1D, CC-M1D, CC-2D, OF-C1D, OF-M1D, OF-2D. In the next section, we compare the respiratory signals obtained by six core algorithms in two different processing frameworks using the phantom benchmark.

\section{Evaluation frameworks and metrics}
The six core algorithms are embedded in a framework for benchmarking. In this section, we describe the front and end processing present in the framework as well as the three metrics used in the performance analysis. There are two flavors in the front processing: fixed- and auto-RoI. For fair comparison, framework settings were kept identical when running different core algorithms.


\subsection{Respiratory signal generation}
Motion estimation core algorithm generates pixel velocities between two video frames. To create a long-term respiratory signal over the video sequence, we first concatenate the velocity of pixels in the vertical direction (i.e. assumed respiratory direction) measured between frame pairs: 
\begin{equation}
\label{eq:10}
               \mathbf{Y} = \{v_{y1}, v_{y2},...,v_{yn}\}.
\end{equation}
Then we use cumulative summation (i.e.\ discrete-time integration) to convert the velocity signal to a respiratory signal:
\begin{equation}
\label{eq:11}
	S_i = \sum_{1}^{i} Y_i,
\end{equation} 
where $Y_i$ is the $i$-th element of $\mathbf{Y}$. Based on $\mathbf{S}$ (the time series of $S_i $), a simple peak detector\footnote{The basic Matlab function $\mathbf{findpeaks(\cdot)}$ with default settings is used to detect peaks in a time signal.} is applied to find peaks in the raw respiratory signal in the time domain, i.e. peaks in our notation denote inhaling. We emphasize that no post-processing (signal smoothing or filtering) is used in order to reveal the true/bare performance of motion estimation, e.g. signal characteristics and noisiness are preserved. 

\subsection{Fixed-RoI framework} We first use the simplest fixed-RoI framework to investigate the bare performance of core algorithms, where an image patch (e.g. $24\times60$ pixels of $360\times480$ pixels frame resolution) targeting the respiratory RoI (e.g. phantom location) was manually selected for respiratory signal extraction (see examples of fixed-RoI in Fig.~\ref{fig:maps_kernel}). When computing $v_y$ between two frames, the frame distance is set to 1 frame (67\,ms for 15 fps camera) and 3 frames (200\,ms for 15 fps camera), respectively, for six core algorithms. Such a comparison is intended for understanding the motion sensitivity of core algorithms in boundary conditions, though the 200\,ms interval is a more usual setting~\cite{Bartula2013Procor}. 

\subsection{Auto-RoI framework} Next, we plugged six core algorithms in an end-to-end respiratory signal measurement framework that performs automatic RoI detection~\cite{Rocque2016Procor}, where the core algorithms are compared on a system-level. As depicted in Fig.~\ref{fig:autoroi}, the auto-RoI framework has three main steps: (i) dense segmentation that segments the input video into multiple blocks; (ii) respiratory signal extraction per block, where six core algorithms are used independently; and (iii) SNR based RoI selection that selects the blocks with clean respiratory signals as the RoI and combines respiratory signals from the RoI into a final output. The benchmark setting is: the block segmentation is set to $12\times30$\,pixels per block with half overlap on the vertical direction (for $360\times480$\,pixels frame resolution). The frame interval for core algorithms is set to 3 frames (200\,ms for 15 fps camera) by default. The signal SNR is computed by a temporal sliding window with 30\,s length, as the ratio of spectrum peak (detected inside the respiratory band [5, 60]~bpm) and total spectrum energy in the frequency domain. 




\begin{figure*}[t!]
\centering
\includegraphics[trim=0cm 0cm 0.5cm 0cm, width=\linewidth]{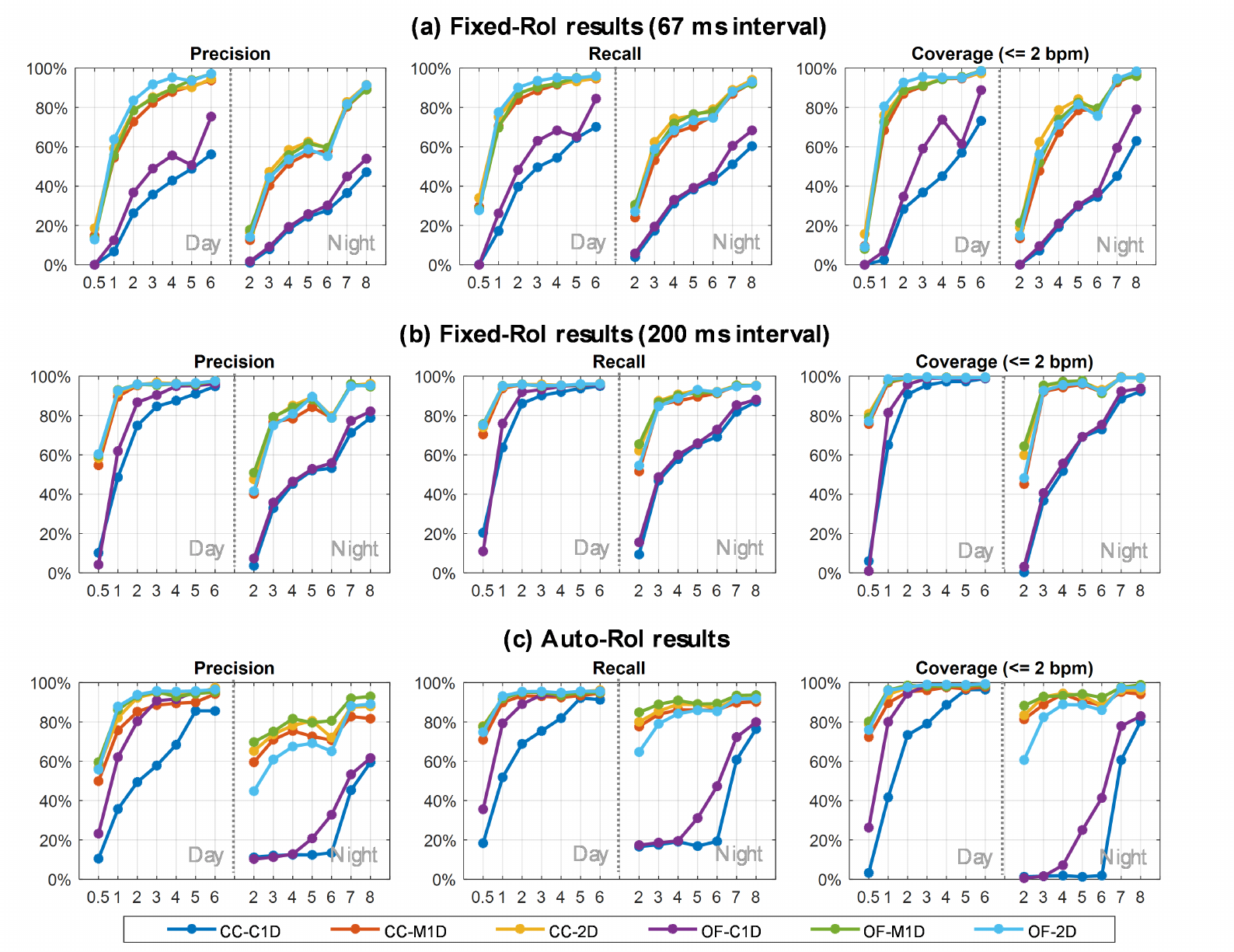}
\vspace*{-0.6cm}
\caption{The performance curves of six core algorithms with different evaluation frameworks and settings in the phantom benchmark: (a) fixed-RoI (67\,ms frame interval); (b) fixed-RoI (200\,ms frame interval), and (c) auto-RoI (200\,ms frame interval by default). The unit of x-axis is mm (denoting amplitude of phantom motion).}
\label{fig:curves}
\vspace*{-0.2cm}
\end{figure*}

\subsection{Evaluation metrics} The breath-to-breath accuracy between the phantom signal (ground-truth) and camera respiratory signal is evaluated based on the detected respiratory peaks in the time domain. For each peak in the reference signal, we set a tolerance window centered around the peak. The window length is 50\% of inter-beat interval w.r.t. its preceding and proceeding peaks, adapted to the instantaneous rate. If a single respiratory peak is detected from the camera signal within the tolerance window, it is counted as a valid measurement. The instantaneous rate associated with this peak (i.e. time instant) is taken as the mean of the rates derived from the Inter-Beat Interval (IBI) with the previous and next peaks (if three consecutive peaks are detected). We use three metrics to quantify the breath-to-breath accuracy: 
\begin{itemize} 
	\item Precision: percentage of valid camera measurement w.r.t. the total number of detected camera peaks (e.g. accuracy).
	\item Recall:  percentage of valid camera measurement w.r.t.  the total number of reference peaks (e.g. retrieval rate).
	\item Coverage ($\le$ 2\,bmp): percentage of instantaneous camera rates that have a deviation $\le$ 2\,bpm w.r.t. the reference instantaneous rate. 
\end{itemize}
A core algorithm that has higher values for these three metrics is considered to have better performance.

As an additional performance measure targeting indicators for the auto-RoI framework, the RoI correspondence is introduced. The correspondence gives the percentage of blocks chosen by the auto-RoI detection method that were also used in the fixed-RoI and therefore agreeing with an expert opinion. Since the selection in auto-RoI is frame-based, the correspondence is a time-varying measure and is provided in terms of mean and standard deviation.

\begin{figure*}[t!]
\centering
\begin{minipage}[t!]{\linewidth}
\includegraphics[trim=0cm 0cm 0cm 0cm, width=\linewidth]{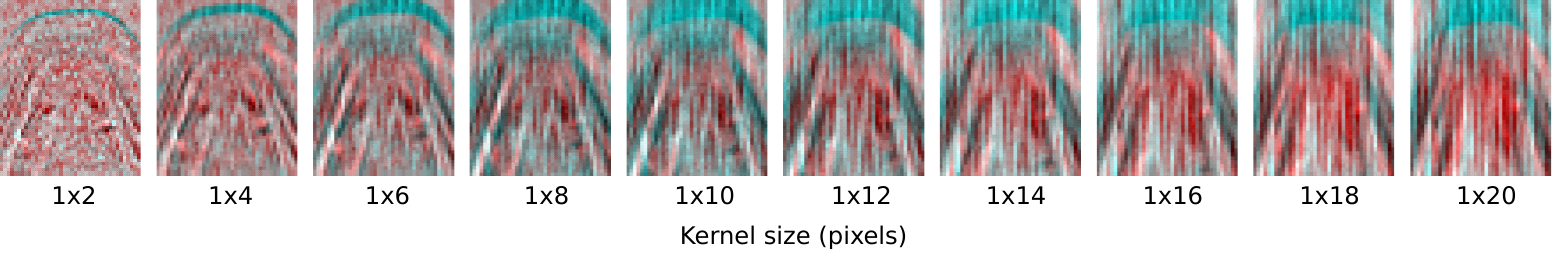}
\vspace*{-0.8cm}
\caption{Example of spatial gradients (red channel) and temporal gradients (green channel) obtained by OF-M1D with different kernel sizes, ranged from $1\times2$ to $1\times20$ pixels. The visualization is based on the video of day (0.5\,mm). With the increase of kernel size, the spatio-temporal gradients are more blurred.}
\label{fig:maps_kernel}
\end{minipage}

\vspace*{0.2cm}

\begin{minipage}[t!]{\linewidth}
\centering
\includegraphics[trim=0cm 0cm 1cm 0cm, width=\linewidth]{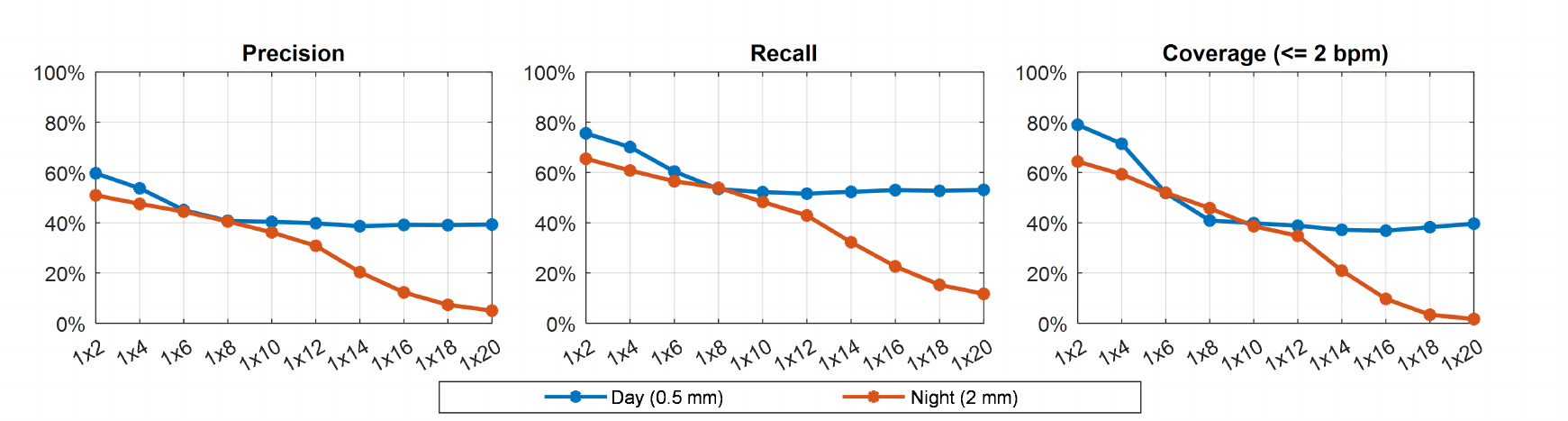}
\vspace*{-0.6cm}
\caption{The performance curves of OF-M1D obtained in the two most challenging sessions (day (0.5\,mm) and night (2\,mm)) with different kernel sizes in the fixed-RoI framework (200\,ms interval).}
\label{fig:curves_kernel}
\vspace*{-0.2cm}
\end{minipage}

\end{figure*}

\section{Results and discussion}
In this section, we discuss the performance of core algorithms in two evaluation frameworks. 

\subsection{Evaluations in the fixed-RoI framework}
Fig.~\ref{fig:curves} (a)-(b) shows the session results obtained in the fixed-RoI framework with two different frame interval settings (67\,ms and 200\,ms). From these figures it is immediately clear that all algorithms perform better on the day-time than on the night-time recordings. The sensor noise in low-light conditions of the night category dominates the pixel changes and interferes with the measurement. With the increase of the motion level, all methods show improved results in each category. When increasing the frame interval for motion estimation (i.e. from 67\,ms to 200\,ms), all methods show clear improvements, which is expected as subtle motion corresponds to larger pixel displacement at longer temporal distances. 

Comparing the six core algorithms, we see that the options with the combined 1D profile (C1D) have clearly worse performance than others (see CC-C1D and OF-C1D in Fig.~\ref{fig:curves} (a)-(b)). This implies that C1D is not a robust spatial representation for motion estimation; neither for cross-correlation nor optical flow. The compression of all image pixels on a single direction combines moving pixels and stationary pixels upfront the motion estimation, which reduces the motion sensitivity. A strategy that emphasizes the motion sensitivity should be first estimating the local velocity per image column (as a local profile) and then combining local velocities into a global velocity representation, where spatial redundancy property of an image sensor is exploited, such as the M1D profile.  The other four algorithmic combinations (CC-M1D, CC-2D, OF-M1D and OF-2D)  perform comparably. 


To verify our hypothesis that small kernels are essential for OF-based methods to attain high motion-sensitivity, we took OF-M1D as an example and executed a set of experiments by changing the spatial kernel size from $1\times$2 (default) to $1\times20$ pixels. The experiment is conducted on the most challenging video session per category: day (0.5\,mm) and night (2\,mm). The spatial gradient maps and temporal gradient maps of OF-M1D are exemplified in Fig.~\ref{fig:maps_kernel}. It can be seen that the gradients measured by large kernels are more blurred than the ones measured by small kernels. They are more sensitive to the (global) changes at larger scales than neighboring pixel changes. Fig.~\ref{fig:curves_kernel} shows the evaluation results of different kernels. With the increase of the kernel size, OF-M1D has consistent quality drops in both video sessions and the degradation is clear. By changing the kernel size from $1\times2$ to $1\times20$ pixels, the coverage is reduced from 80\% to 40\% for day (0.5\,mm), from 60\% to less than 5\% for night (2\,mm). This suggests that smaller kernels are indeed preferred for OF-based respiratory motion extraction algorithms in order to be sensitive to small motions occurring between neighboring pixels. We stress that though the use of small kernels may increase the method's immunity to large motions, but we would not call it ``motion robustness''. If large motion is a global motion that influences the full image, it will pollute the measurement of local motions in the foreground RoI.

\subsection{Evaluations in the auto-RoI framework}

Fig.~\ref{fig:curves} (c) shows the session results obtained in the auto-RoI framework (end-to-end processing). Obviously, the methods using C1D profile is worse than others, which confirms our observation in the fixed-RoI experiment. The methods using either M1D or 2D profile have a rather similar performance, i.e. OF-M1D seems to have slightly better performance, followed by CC-2D, CC-M1D and OF-2D. As a numerical comparison, we show the statistical values (mean and standard deviation over sessions) of six core algorithms in Table II. The best performance numbers are dominantly obtained by OF-M1D and for the night condition all three performance metrics attain their maximum for this algorithm. It has an averaged precision, recall and coverage of 
88.1\%, 91.8\% and 95.5\% in the day category, and 81.7\%, 90.0\% and 93.9\% in the night category. Since the vertical direction is clearly the dominant motion direction, there is little room for performance improvement by exchanging OF-M1D for OF-2D. In fact we find the opposite: in the night-time simulations, the 2D approach performs less presumably because the freedom to estimate the horizontal motion introduces additional measurement uncertainty. The OF-2D that estimates both the vertical and horizontal motions on a 2D plane will absorb a portion of temporal intensity changes into the horizontal direction that is orthogonal to the respiratory direction, which reduces the sensitivity as compared to OF-M1D that only estimates velocities on the vertical direction. In OF-M1D, all temporal image intensity changes are translated into the vertical velocity of the profile.

In line with the above, one may argue that OF-M1D attains better sensitivity than OF-2D for the following reason. Due to the characters of exhaling and inhaling, respiratory motion is not equally strong in all directions. There is always a direction where the respiratory energy is stronger. To maximize the sensitivity of a motion algorithm like optical flow, we may use all temporal intensity changes to estimate/regress the velocity on a single direction with pre-assumed larger respiratory energy instead of spreading the estimation over different directions. The main respiratory direction can be determined based on the monitoring scenario or setup (e.g. sleep monitoring, triage screening). Once the setup is fixed, the assumption will be stable.

\begin{figure*}[t!]
\centering
\includegraphics[trim=0cm 0cm 0.5cm 0cm, width=\linewidth]{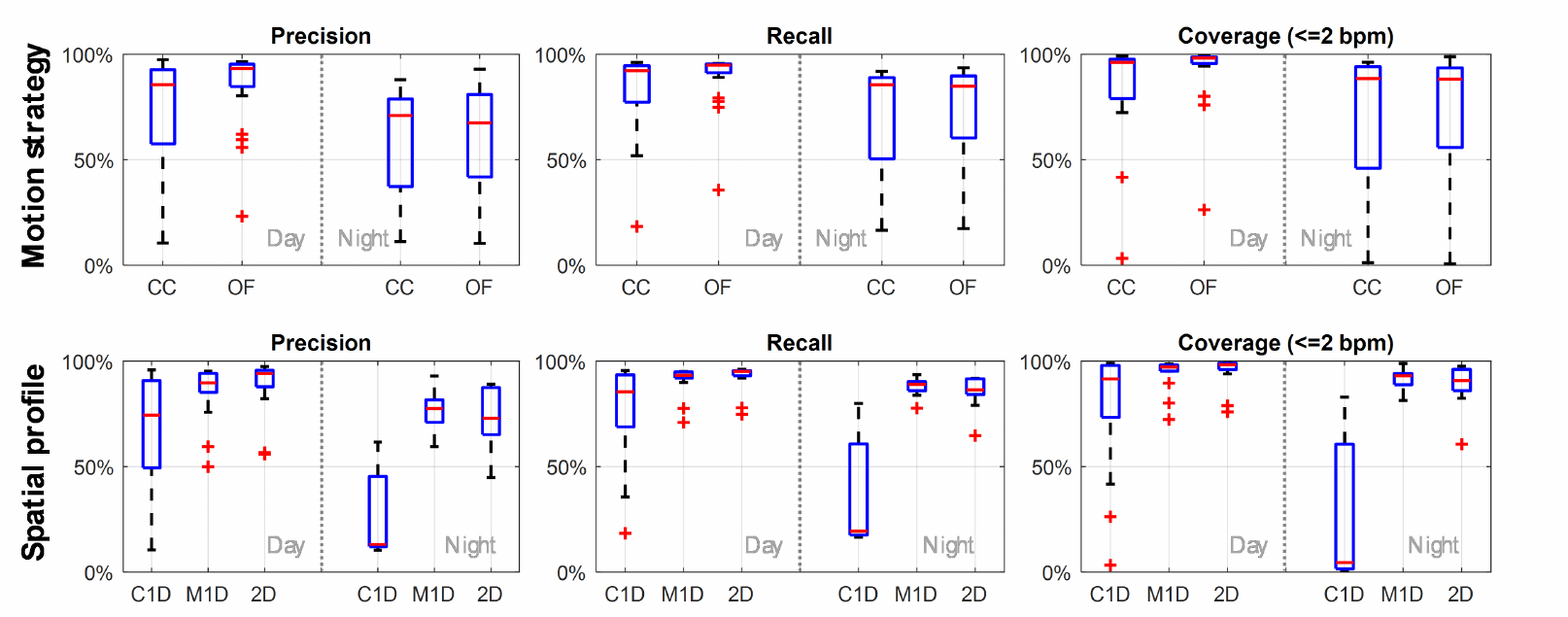}
\vspace*{-0.8cm}
\caption{The boxplot of benchmark results in terms of motion strategy (CC and OF) and spatial profile (C1D, M1D and 2D). The median values are indicated by horizontal bars inside the boxes, the quartile range by boxes, the full range by whiskers.}
\label{fig:bars}
\vspace*{-0.2cm}
\end{figure*}

To get more insights into the components of core respiratory algorithms, we show the boxplot of all benchmark results in terms of motion estimation strategies (CC and OF) and spatial representations (C1D, M1D and 2D) in Fig.~\ref{fig:bars}. OF is generally better than CC in the day category, while in the night category they are rather similar. Regarding the spatial representation, M1D and 2D profiles are considerably better than C1D. M1D is slightly better than 2D, but the difference is considered not significant.

\begin{table*}[!t]
\small	
\centering
\renewcommand{\arraystretch}{1.5}
\label{tab:4}
\caption{Statistical values (mean and standard deviation) of six core algorithms obtained in the auto-RoI framework.}
\vspace{-0.2cm}
\begin{tabular}{c|c|ccc|ccc}
\hline
\hline
\multirow{2}{*}{Evaluation metric} & \multirow{2}{*}{Category} & \multicolumn{3}{c|}{Cross-Correlation (CC)} & \multicolumn{3}{c}{Optical Flow (OF)}\\
\cline{3-8}
& & CC-C1D & CC-M1D & CC-2D & OF-C1D & OF-M1D & OF-2D \\
\hline
\multirow{2}{*}{Precision (\%)} 	& Day  	& 56.2 [27.2]&  81.9 [15.2]&       87.4 [14.4]&       77.0 [26.5]&       88.1 [13.0]&      \textbf{88.7 [14.8]} \\
							& Night  	& 23.8 [20.0]&  73.4 [7.81]&       77.9 [8.29]&       29.0 [21.1]&       \textbf{81.7 [8.41]} &      69.2 [15.5]\\
\hline
\multirow{2}{*}{Recall (\%)} 	& Day  	& 68.6      [26.1]&      89.7 [8.39]&       \textbf{92.3 [6.49]} &      83.2     [21.7]&        91.8    [6.39]&     92.1  [7.69] \\
							& Night  	& 32.4      [25.2]&      85.7 [4.19]&       87.7      [4.08]&      40.9     [26.3]&        \textbf{90.0    [3.02]}&     83.3   [9.3]\\
\hline
\multirow{2}{*}{Coverage (\%)} 	& Day  	& 68.4      [34.3]&      92.2 [9.19]&       95.0      [7.31]&       85.0    [26.8]&     \textbf{95.5      [6.81]} &     95.1    [8.53] \\
							& Night  	& 21.3      [34.1]&      90.4 [4.83]&       92.1      [4.64]&      33.8     [35.0]&     \textbf{93.9      [3.51]} &     85.9    [12.5]\\
\hline
\hline
\end{tabular}
\begin{tablenotes}
\footnotesize
\item * \textbf{boldface} entry denotes the best combination per row. Numbers outside and inside the brackets denote mean and standard deviation.
\end{tablenotes}
\end{table*}

\begin{figure*}[t!]
\centering
\begin{minipage}[t!]{\linewidth}
\includegraphics[trim=0cm 0cm 0cm 0cm, width=\linewidth]{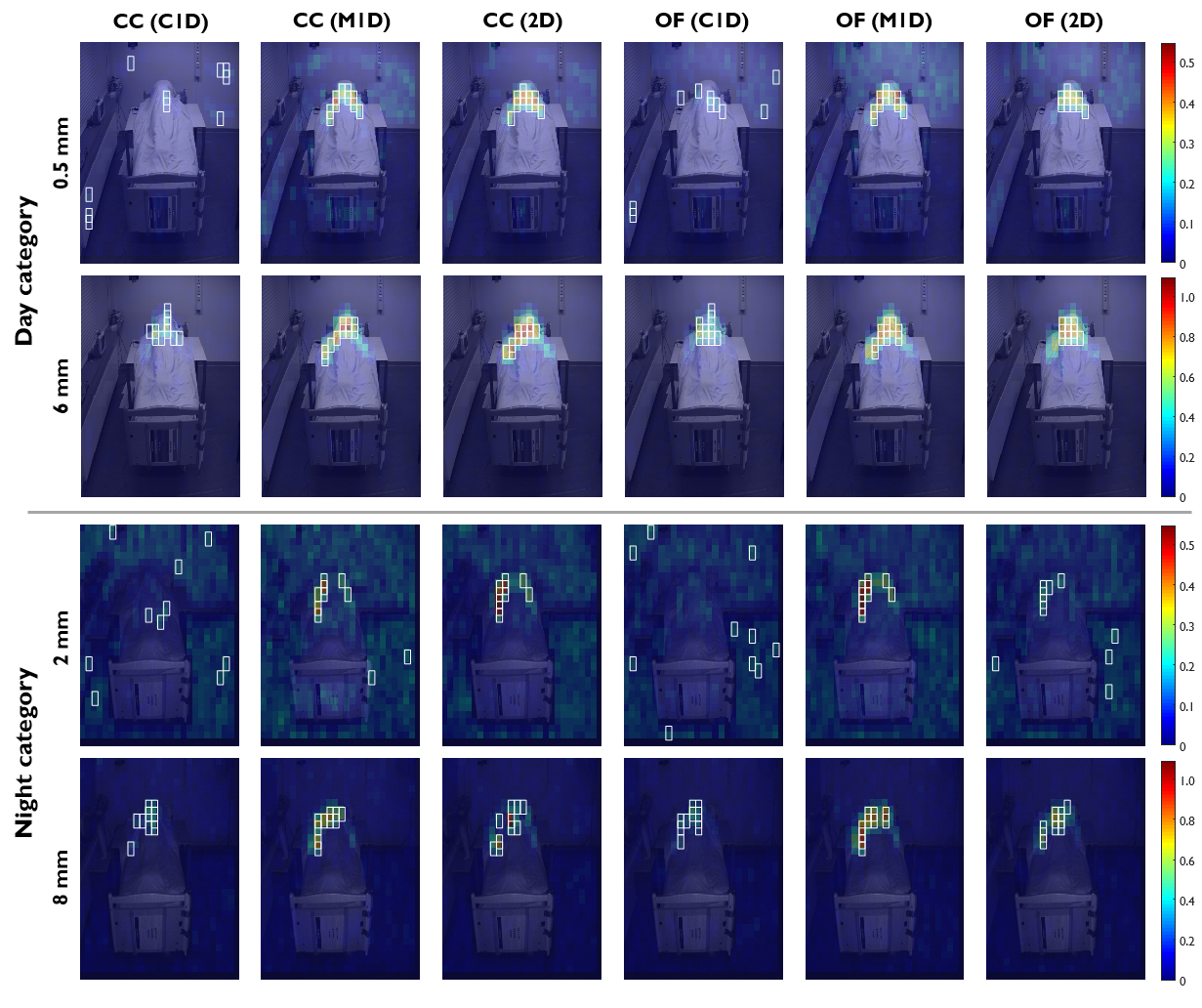}
\vspace*{-0.8cm}
\caption{Examples of detected RoIs in the auto-RoI framework with six core algorithms. Sessions with minimum and maximum signal excursions are used for demonstration. The color scale denotes the range of SNR values calculated from respiratory signals in segmented blocks.}
\label{fig:roi}
\vspace*{0.2cm}

\includegraphics[trim=0cm 0cm 0.5cm 0cm, width=\linewidth]{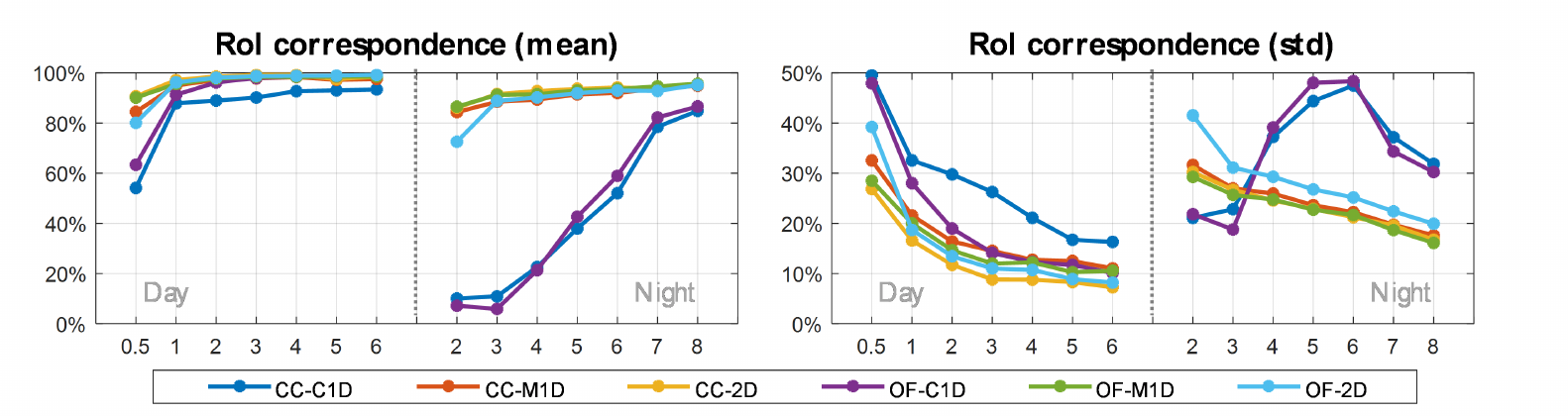}
\vspace*{-0.6cm}
\caption{The correspondence of auto-RoI detection for six core algorithms in day and night categories. Higher mean values denote more accurate RoI detection, while lower standard deviation values denote more stable RoI detection.}
\label{fig:roi_eval}
\vspace*{-0.2cm}
\end{minipage}

\end{figure*}


Another dimension to assess the performance of core algorithms is via the RoI detection (e.g. RoI correspondence), because auto-RoI detection is based on the SNR of respiratory signals. The block segments showing cleaner respiratory signals are more likely to be selected as the RoI. Fig.~\ref{fig:roi} exemplifies the RoIs detected by six core algorithms in the most challenging session (with smallest motion level) and the simplest session (with largest motion level) per category. In the day category (0.5\,mm session), only CC-C1D and OF-C1D cannot find the RoI, which explains their poor performance in our benchmark, i.e. the RoI selection was wrong. In the night category (2\,mm session), CC-2D and OF-M1D have the best RoI-detection performance, while the rest are more or less suffering from false positives (i.e. the selected RoI blocks are more spreading and less focused on the phantom). None of the algorithms has a problem in finding the RoI in the simplest session (with the largest excursion) in both categories. Fig.~\ref{fig:roi_eval} shows the correspondence to the fixed-RoI for six core algorithms in the benchmark dataset. Better core algorithms  show better RoI-detection performance, which means they have higher overlap values with the ground-truth and the overlap is also more time-stable (less jitter). The conclusion drawn from the RoI correspondence is consistent with the findings based on the performance measures. 

We summarize our insights as follows. To create a motion-sensitive algorithm, the choice for the spatial representation (profile) is highly important. We recommend to use the spatial redundancy of image pixel sensors to estimate local motions before generating a global motion representation. This is better than first combining image pixels into a global spatial representation and then estimating the global motion because this essentially presumes motion of rigid objects which is in reality hardly ever the case. To estimate the velocity between profiles from subsequent video frames, cross-correlation and optical flow do not show significant difference in an overall sense (see Fig.~\ref{fig:bars}), but optical flow is more sensitive as shown by the increased numbers for all performance criteria, especially  with larger differences (relative to the standard deviation) for the night-time performance (see Table~II). This supports the notion that use of small kernels (with small receptive field as in OF) is important. We mention that we did not use off-the-shelf OF algorithms as these typically include additional operations intended for other purposes (e.g. object tracking, large motion estimation) and we found them to perform poorly for the case of respiration monitoring. Motion sensitivity benefits from using the assumption of a dominant motion direction; introducing an extra freedom to allow for 2D displacements did not increase performance and requires more computing power. The summarized insights are incorporated in one of the six benchmarked core algorithms: OF-M1D (for reproducibility purpose, we provide the pseudo-code in Algorithm 1), which is proved to be a highly-sensitive algorithm in our benchmark. The essential steps are simple to understand and implement (in a few lines of Matlab code), and its performance is easy to reproduce.

\begin{algorithm}[!t]
\renewcommand{\algorithmicrequire}{\textbf{Input:}}
\renewcommand{\algorithmicensure}{\textbf{Output:}}
\renewcommand{\algorithmicprint}{\textbf{Initialize:}}
\caption{Highly-sensitive respiratory signal extraction}\label{alg:1}
	\begin{algorithmic}[1] 
	\REQUIRE A video sequence with $N$ frames. 
	\PRINT A manually or automatically defined $\mathbf{RoI}$; $\Delta t=3$ (e.g. for 20 fps camera); $\mathbf{R} = 0;$ 
  	\FOR{$i = 1,...,N-\Delta t$}           
		\STATE $\mathbf{I_i} \leftarrow \mathbf{frame_i}(\mathbf{RoI})$; $\mathbf{I_{i+\Delta t}} \leftarrow \mathbf{frame_{i+\Delta t}}(\mathbf{RoI})$;
		\STATE $\mathbf{\bar{I}_i} = \mathbf{I_i} ./ \mathsf{mean}(\mathbf{I_i},1); \rightarrow$ per column DC-normalization
		\STATE $\mathbf{\bar{I}_{i+\Delta t}} = \mathbf{I_{i+\Delta t}} ./ \mathsf{mean}(\mathbf{I_{i+\Delta t}},1);$
		\STATE $\mathbf{{Dy}} = \mathsf{conv2}(\mathbf{\bar{I}_i}, [-1;1], '\mathsf{valid}');$
		\STATE $\mathbf{{Dt}} = \mathsf{conv2}(\mathbf{\bar{I}_i}-\mathbf{\bar{I}_{i+\Delta t}}, [1;1], '\mathsf{valid}');$
		\STATE $\mathbf{R_{i+1}} = \mathsf{sum}(\mathbf{{Dy}(:)}.*\mathbf{{Dt}(:)})/\mathsf{sum}(\mathbf{{Dy}(:)}.*\mathbf{{Dy}(:)});$
    	\ENDFOR
    	\STATE $\mathbf{Resp} = \mathsf{cumsum}(\mathbf{R}, 2);\rightarrow$ cumulative sum
\ENSURE The respiratory signal $\mathbf{R}$.
\end{algorithmic}
\end{algorithm}
\section{Conclusion}
To increase the insights needed for the development and application of well-functioning respiration monitors, we have made several contributions in this paper. A model was formulated to outline the basic principles currently used in camera-based respiration monitoring. We have made a benchmark using the two core principles, cross-correlation and optical flow, where we used a phantom source to mitigate limitations (full coverage of different rates and signal strengths) imposed by trials involving subjects.  We have sketched the influence of using different spatial profiles. Good performance (coverage above 90\%) was obtained by simple and explainable algorithms and especially the OF-M1D performed well also in the more difficult cases (night-time, auto-RoI).

\section*{Acknowledgment}
The authors would like to thank Mr. Jacob Vasu for creating the benchmark dataset; Mr. Benoit Balmaekers, Mr. Age van Dalfsen, and Mr. Mukul Rocque for the discussions on the topic.



\bibliographystyle{IEEEtran}
\bibliography{bibliography}

\end{document}